\documentclass[conference]{IEEEtran}
\IEEEoverridecommandlockouts
\usepackage{cite}
\usepackage{amsmath,amssymb,amsfonts}
\usepackage{algorithmic}
\usepackage{graphicx}
\usepackage{textcomp}
\usepackage{xcolor}
\usepackage{amssymb}
\usepackage{algorithm}
\usepackage{algorithmic}
\usepackage{amsmath}
\usepackage{graphicx}
\usepackage{multirow} 
\usepackage{tabularx}
\usepackage{booktabs}
\usepackage{hyperref}
\usepackage{float}
\usepackage{mathrsfs}

\def\BibTeX{{\rm B\kern-.05em{\sc i\kern-.025em b}\kern-.08em
    T\kern-.1667em\lower.7ex\hbox{E}\kern-.125emX}}

\begin{document}

\title{MC-DBN: A Deep Belief Network-Based Model for Modality Completion\\
\thanks{{*} These authors contributed equally to this work.}
\thanks{{\dag} Corresponding Author.}
}

\author{
    \IEEEauthorblockN{Zihong Luo$^{a,*,\dag}$, Zheng Tao$^{a,*}$, Yuxuan Huang$^{a,*}$, Kexin He$^b$, Chengzhi Liu$^a$}
    \IEEEauthorblockA{
        $^a$ Xi'an Jiaotong-Liverpool University,
        $^b$ Kean University
    }
    \IEEEauthorblockA{
        \textbf{Email:} \{Zihong.Luo22, Zheng.Tao22, Yuxuan.Huang2002, Chengzhi.Liu21\}@student.xjtlu.edu.cn, hekex@kean.edu.
    }
}

\maketitle

\begin{abstract}

Recent advancements in multi-modal artificial intelligence (AI) have revolutionized the fields of stock market forecasting and heart rate monitoring. Utilizing diverse data sources can substantially improve prediction accuracy. Nonetheless, additional data may not always align with the original dataset. Interpolation methods are commonly utilized for handling missing values in modal data, though they may exhibit limitations in the context of sparse information. Addressing this challenge, we propose a Modality Completion Deep Belief Network-Based Model (MC-DBN). This approach utilizes implicit features of complete data to compensate for gaps between itself and additional incomplete data. It ensures that the enhanced multi-modal data closely aligns with the dynamic nature of the real world to enhance the effectiveness of the model. We conduct evaluations of the MC-DBN model in two datasets from the stock market forecasting and heart rate monitoring domains. Comprehensive experiments showcase the model's capacity to bridge the semantic divide present in multi-modal data, subsequently enhancing its performance. The source code is available at: \href{https://github.com/logan-0623/DBN-generate/}{https://github.com/logan-0623/DBN-generate/}.
\end{abstract}

\begin{IEEEkeywords}
    Multi-Modal, DBN Network, Stock Market Forecasting, Heart Rate Monitoring
\end{IEEEkeywords}

\section{Introduction}


The significant advancements in artificial intelligence and multi-modal technologies have profoundly impacted the fields of stock market prediction and heart rate monitoring. These advanced methodologies integrate various data types\cite{wang2019aggregating}, including numerical, textual, and visual inputs, to provide comprehensive insights into market trends and physiological health indicators. In stock market prediction, the inclusion of intermittent and non-sequential data, such as financial news and policy updates, is crucial. However, these data types are likely to face the risk of incomplete modalities due to missing data. Similarly, heart rate monitoring necessitates the combination of different physiological and environmental data for accurate health predictions, where missing modalities also present challenges. Traditional modal completion techniques, such as linear interpolation methods\cite{wang2007new}, often fail to capture the true trends and variances in such multi-modal data, leading to suboptimal modelling outcomes. Thus, both fields confront the common issue of potential modality incompleteness in their predictive analyses.

To overcome these limitations, our research introduces the MC-DBN model, an innovative solution designed to intelligently impute missing data, capturing the inherent temporal volatility and patterns in both stock market and heart rate data. This approach not only fills gaps in multi-modal datasets but effectively addresses gaps in multi-modal datasets in alignment with the temporal dynamics of the data, enhancing the robustness and reliability of predictive models. Our work significantly contributes to the fields of stock market forecasting and heart rate monitoring in the following three aspects:
\begin{itemize}
\item\textbf{Innovative Multi-modal Data Integration Framework:}
We devise a multi-modal data integration framework based on modalities with missing information. The core components are a modality completion encoder-decoder framework and an attention fusion module. It employs a multi-head cross-attention mechanism to effectively capture and weight the multi-modal information. Additionally, we utilize operations such as mapping and normalization to efficiently integrate multi-modal data.

\item  \textbf{Enhanced Capability for Data Completion:}

The modal completion encoder-decoder framework integrates deep belief networks (DBN) with attention mechanisms. 
Within the decoder structure, we purposefully select targeted feature extraction architectures for different modal information. This method demonstrates outstanding performance in various applications, including heart rate detection and stock market prediction.

\item \textbf{Cooperative Modal Completion Loss Function:}\\
We design two collaborative loss functions. The first one is a dedicated encoder loss function designed for completing missing modality information. It optimizes the quality of the encoder's data completion by comparing the completed data with the original information. The second is a global loss function constructed based on diverse downstream tasks, aiming to optimize the overall network performance.

\end{itemize}

\section{Related work}

\subsection{Multi-modal Data Integration}


Multi-modal data, which encompasses data from diverse sources and modes such as numerical, categorical, and textual information, plays a crucial role in various domains, including finance \cite{Holtz2023MultimodalTF}. In the financial sector, multi-modal data is instrumental for tasks like risk classification, incident detection, and stock price forecasting \cite{hozhabr2022machine, Buche2023EnhancingPM}. For instance, one study employed a Multi-modal Transformer for risk classification, exploring data augmentation and usage by automatically retrieving news articles \cite{Holtz2023MultimodalTF}. Similarly, in the context of heart rate monitoring, research \cite{s20082186} combined physiological data with environmental and behavioural factors for a comprehensive health analysis. The inclusion of both numerical and categorical data was found to enhance model performance, particularly for risks that are difficult to classify using text data alone \cite{Holtz2023MultimodalTF}. Incorporating a variety of data types has been shown to improve model performance in both areas, especially in scenarios where single-modal data is insufficient. Another study introduced a financial forecasting method utilizing a hybrid module that combines BERT and BiLSTM for mixed information processing \cite{Buche2023EnhancingPM}.

\subsection{Stock Market Forecasting and Heart Rate Monitoring}
The convergence of stock market forecasting and heart rate monitoring in the field of multi-modal data analysis has opened new avenues in both the financial and healthcare sectors.

In stock market forecasting, AI and machine learning algorithms have been increasingly employed to predict market trends and movements. A notable example is the use of Deep Learning models, such as Convolutional Neural Networks (CNNs) and Recurrent Neural Networks (RNNs), to analyze historical price data and market sentiment extracted from news articles and social media.  The work of Mogharet al. (2020) demonstrates the effectiveness of using LSTM networks for predicting stock prices based on historical data \cite{moghar2020stock}. Additionally, the integration of sentiment analysis from financial news, as explored by Agarwal et al. (2020), provides a more holistic approach to forecasting \cite{agarwal2020sentiment}.

Concurrently, heart rate monitoring has seen significant advancements with the application of AI. Heart rate data, collected through wearable devices, is analyzed using sophisticated algorithms to detect anomalies and predict potential health risks. A study by Bertsimas et al. (2021) highlights the use of machine learning techniques for real-time heart rate monitoring and anomaly detection \cite{bertsimas2021machine}. Moreover, the combination of heart rate data with other physiological parameters for comprehensive health assessment is explored in the research by Hussain and colleagues (2020) \cite{hussain2020detecting}.

\vspace{-5pt}

\subsection{Methods for handling missing values}

Handling missing values is essential in financial analysis and heart rate monitoring. Missing data, which can result from various causes like incomplete data entry or equipment malfunctions, leads to biases and reduced statistical power, affecting the validity of conclusions \cite{kang2013prevention, donders2006gentle}. The three primary types of missing data are missing completely at random (MCAR), missing at random (MAR), and missing not at random (MNAR) \cite{donders2006gentle}.

Addressing missing data involves methods such as imputation or data removal \cite{bennett2001can}. Traditional imputation techniques include the Last Observation Carried Forward (LOCF) and the Next Observation Carried Backward (NOCB), which may introduce biases \cite{lachin2016fallacies, engels2003imputation}. Advanced techniques like rolling statistics and interpolation methods offer more sophistication but can struggle with irregular patterns in financial and heart rate data \cite{broadstock2020integration, arun2013comparative}.

PCA-based methods like bPCA and llsPCA have been effective in financial time series and can be adapted for heart rate data \cite{schneider2010benzene, john2019imputation}. Deep learning-based imputation, including Multi-Layer Perceptrons (MLP), is emerging for long-term missing value estimation in both fields \cite{kruse2022multi}.

\section{Methodology}

\begin{figure*}
    \centering
    \vspace{-25pt}
    \includegraphics[width=\textwidth]{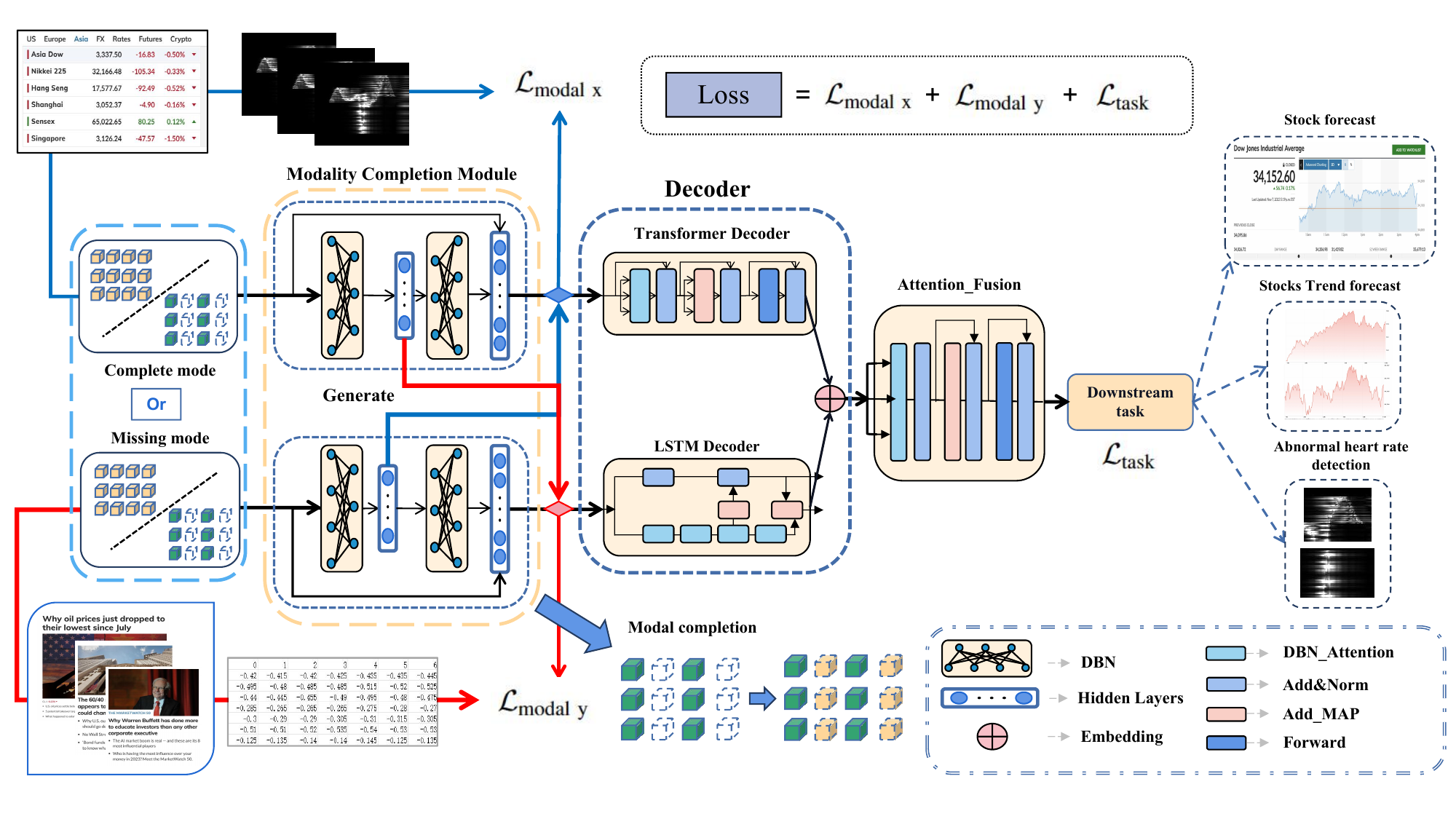}
    \vspace{-30pt}
    \caption{The network comprises three main components. Initially, the modality completion module, leveraging a Deep Belief Network (DBN), identifies and fills absent modal data. Secondly, the modality feature extraction module employs Transformer\cite{lin2022survey} and LSTM architectures\cite{yu2019review} for proficient feature extraction. Lastly, a fusion module with an attention mechanism integrates these features. Subsequently, the downstream network generates tailored predictions for multi-modal sequence data. This framework adeptly manages scenarios involving incomplete multi-modal data, which is a common situation in datasets like stock or heart rate data.}
    \vspace{-12pt}
    \label{fig:DBN}
\end{figure*}

\subsection{RBM-based Latent Representation Learning}
To enhance and elaborate on our methodology described, it's essential to give a comprehensible explanation of how Restricted Boltzmann Machines (RBMs) work and then discuss how they are applied in the given context. 

RBMs play a crucial role in unsupervised learning by identifying the probabilistic characteristics of input data. They form a bipartite graph consisting of visible units (representing input data) and hidden units (capturing latent features). The 'restricted' nature of RBMs means that there are no connections within either the visible or hidden layers, only between them.

The learning process in RBMs involves adjusting the weights and biases to minimize the reconstruction error of the input data. This is typically achieved using training algorithms like Contrastive Divergence, which approximates the gradient of the log-likelihood.
\begin{equation}
p(\mathbf{h}|\mathbf{v}) = \prod_{j} p(h_j | \mathbf{v}), \quad p(\mathbf{v}|\mathbf{h}) = \prod_{i} p(v_i | \mathbf{h})
\end{equation}
In this context, \( \mathbf{h} \) and \( \mathbf{v} \) represent the hidden and visible units, respectively.

The integration of RBM in this manner allows the model to capture and utilize deep non-linear relationships within the data, enhancing feature representation and supporting the generation of new synthetic data samples consistent with the learned distribution.

Deep Belief Network (DBN) is a sophisticated deep learning architecture featuring an input layer, numerous stacked layers housing random variables, and an output layer. The optimization process initiates with a meticulous layer-by-layer training approach for each Restricted Boltzmann Machine (RBM) network, advancing systematically from the lowermost layer to the uppermost. Following this, the entire network undergoes refinement through fine-tuning, a process facilitated by the backpropagation algorithm.

\subsection{Modal Completion Encoder-Decoder Framework}
\begin{figure}
    \label{fig:stock_market}
    \centering
    \includegraphics[width=0.481\textwidth]{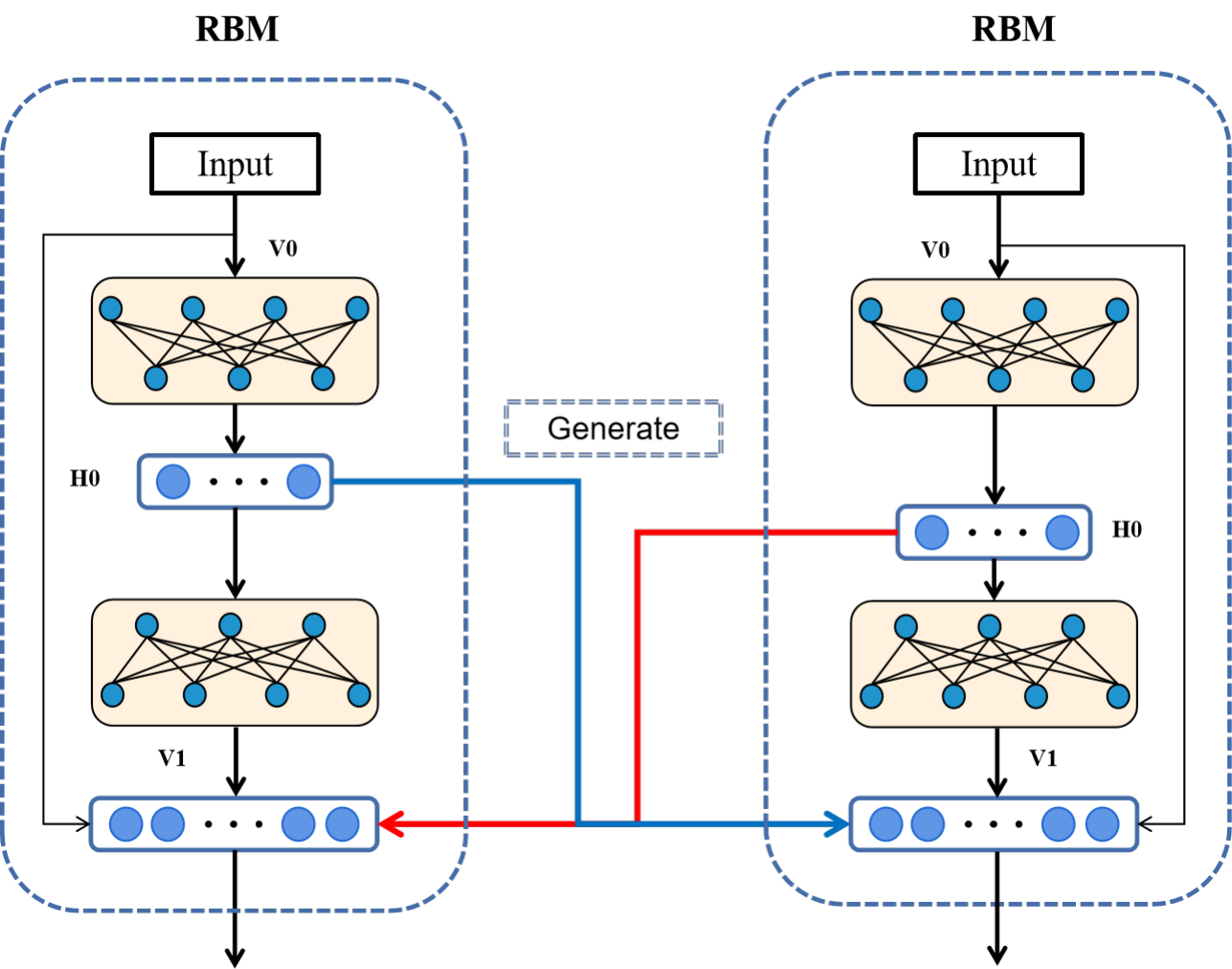}
    \caption{For our modality completion mechanism, the algorithm employs MC-RBMs as probabilistic generative models to capture the latent representations of input data. The quality of the completed modality is optimized using Mean Squared Error (MSE) loss concerning the original modality features, enhancing the effectiveness of modal generation. The features sampled post-completion are then subject to a residual connection with the original features undergoing convolution operations, facilitating further processing and analysis.}
    \vspace{-20pt}
\end{figure}

We propose a modal completion framework based on an encoder-decoder structure designed to effectively address missing modality information (shown in Figure\ref{fig:DBN}). Employing a reverse sampling approach, we leverage the hidden features of a known modality to infer and reconstruct missing data in another modality. This innovative method not only fills gaps in the dataset but also significantly enhances the model's multimodal processing capabilities. Furthermore, we introduce an attention mechanism to perceptually complete modality information from a global perspective. The detailed schematic diagram of the modal completion structure is illustrated in Figure \ref{fig:stock_market}. The workflow of the modal completion framework is outlined in  Algorithm\ref{alg:your_algorithm}.

\begin{algorithm}
    \label{alg:your_algorithm}
    \caption{Modal Completion Encoder-Decoder Module}
    \begin{algorithmic}
        \REQUIRE Data of different modalities $I_{\text{x}}$, $I_{\text{y}}$ \\
        \textbf{Modal Completion Encoder:}
        \STATE $W_{\text{attn}} = \text{Self-Attention}(I_{\text{x,y}})$
        \STATE $Attn_{\text{x,y}} = \text{Softmax}\left(I_{\text{x,y}} \odot W_{\text{attn}}\right)$
        \STATE $\hat{H}_{\text{x,y}} = \text{MC-RBM}_{\text{hidden}
        }(I_{\text{x,y}}, Attn_{\text{x, y}})$\\
        \textbf{Modal Completion Decoder:}
        \IF{$I_x$ is incomplete}
            \STATE $G_{\text{x}} = \text{MC-RBM}_{\text{x-complete}}(\hat{H}_y) \rightarrow \mathcal{L}_{\text{modal x}}(I_x, G_x)$
        \ELSIF{$I_y$ is incomplete}
            \STATE $G_{\text{y}} = \text{MC-RBM}_{\text{y-complete}}(\hat{H}_x) \rightarrow \mathcal{L}_{\text{modal y}}(I_y, G_y)$
        \ENDIF
    
        \STATE $MC_{\text{x}} = \text{Decoder\_Transformer}(G_x)$
        \STATE $MC_{\text{y}} = \text{Decoder\_LSTM}(G_y)$
        
        \RETURN $MC_{\text{x}}$, $MC_{\text{y}}$
    \end{algorithmic}
\end{algorithm}
In the modal completion encoder sub-module, we employ an attention mechanism to process two modal data, denoted as $I_x$ and $I_y$, obtaining corresponding attention weights $W_{attn}$. By appropriately balancing and weighting these attention-weighted representations with the original modal inputs, we obtain a novel attentional input $Attn_{x,y}$. This ensures a global awareness of the underlying modal information. By transforming the attention input into the form of a  Modal Completion Restricted Boltzmann Machine (MC-RBM), we obtain the initial hidden state $h$. This transformation is achieved using a weight matrix $\mathbf{W}$ and a bias vector $\mathbf{b}_h$, with the Sigmoid function $\sigma$ serving as the non-linear transformation.

The hidden state $h$ undergoes a further Bernoulli sampling process, yielding a set of sampled binary hidden states $h_{\text{sampled}}$. This step introduces randomness, simulating the probabilistic nature of the hidden layer in the MC-RBM. The state of each hidden unit is determined by the corresponding unit's probability distribution, and it is barbarized through Bernoulli sampling, forming the final hidden state.
\vspace{-7pt}
\begin{equation}
\mathbf{h}_{sampled} = \text{Bernoulli}\biggl( \sigma(\mathbf{W} \odot Attn_{x, y} + \mathbf{b}_h)\biggr)
\end{equation}
Finally, through the process of MC-RBM on the hidden layer information, a re-modelling of the latent representation of multi-modal data is achieved to capture latent patterns and correlations within the data. In this process, a supplementary operation of cross-modal data mapping is introduced to ensure effective information completion in case of missing data in any modality. Through this transformation, the model can effectively fill in missing modality information,  enhancing the overall data integrity and information representation capability.

In selecting the decoder for our framework, we opted for a combination of Long Short-Term Memory (LSTM)\cite{yu2019review} and Transformer networks\cite{lin2022survey}, recognizing their distinct strengths in handling sequential data and long-term dependencies. LSTM excels at capturing temporal relationships in data, while the Transformer, with its self-attention mechanism, performs exceptionally well in scenarios where input data has complex interdependencies. The combination of these two network structures is particularly suitable for applications like stock prediction and heart rate monitoring, which require handling complex data with long-term dependencies. 

\subsection{Attention Fusion Module} 


The outputs from independently processed modalities are integrated into the fusion module, which transforms raw data from each modality into a comprehensive and nuanced representation. This fusion module is composed of attention layers, mapping layers, and normalization mechanisms. The key component, \( Attention_{dbn} \), globally assesses attention weights across both complete and missing data modalities.

The multi-head attention mechanism in this module selectively concentrates on information from various representation subspaces. It utilizes query (\( Q \)) and key (\( K \)) matrices for complete modality information, and a value (\( V \)) matrix for missing modality data. These matrices are represented as \( Q, K \in \mathbb{R}^{d_q \times k} \) and \( V \in \mathbb{R}^{d_v \times k} \), respectively. The embedding matrices \( W_i^{Q, K, V} \) correspond to \( K, Q, V \), enhancing the fusion module's capacity to utilize both complete and missing modality information effectively.

The attention mechanism is mathematically represented as:
\begin{equation}
Attention_{dbn} = \text{Softmax} \left(\frac{QK^T}{\sqrt{d_k}}\right) \times \text{Norm} (V)
\end{equation}

For each head \( i \) in the multi-head attention mechanism, the process is:
\begin{equation}
\text{Multi\_attn}_i = \text{Concat}\bigl[Attention_{dbn} (Q_i, K_i, V_i) \times W_i^{Q,K,V}\bigr]
\end{equation}

The fusion of feature information is further refined through mapping and normalization layers, forming the \( Fusion \) output. Notably, a decoder supplements this process with a stochastic mechanism, integrating both complete and missing information. This integration significantly enriches the fusion feature's informational depth and relevance.

\begin{equation}
Fusion = \text{Map}\biggl(\text{Norm}(\text{Multi\_attn})\biggr) \oplus \text{decoder}(MC_{x,y})
\end{equation}

\subsection{Multiple loss function design}
Our model employs two Mean Squared Error-based loss functions to optimize and enhance performance. One is a modality completion loss function specifically designed for MC-DBN to accurately complete multi-modal data. The other is tailored for different downstream tasks. These loss functions address different aspects and components of the model, ensuring accuracy and efficiency in handling multi-modal data.

\subsubsection{Loss Function for Modality Completion} 
The first loss function is dedicated to optimizing the generated modality of a specific data type, referred to as Modality A. This function computes the MSE between the generated data of Modality A and the original input data of the same modality.
\begin{equation}
\mathcal{L}_{\text{modal x,y}} = \frac{1}{n} \sum_{i=1}^{n} \biggl(G_{\text{modal x,y}}^{(i)} - I_{\text{modal x,y}}^{(i)}\biggr)^2
\end{equation}
where \( G_{\text{modal x,y}}^{(i)} \) represents the \( i \)-th complete data point for different modalities, and \( I_{\text{model x,y}}^{(i)} \) is the corresponding original data point.

\subsubsection{Loss Function for Specific Downstream Task}
The second loss function is tailored for specific downstream tasks, such as classification, regression, or other predictive modelling tasks. This function is designed to optimize the model's performance in these tasks and is chosen based on the specific requirements of the application.

\vspace{-7pt}
\begin{equation}
\mathcal{L}_{\text{task}} = -\frac{1}{N} \sum_{i=1}^{N} \sum_{j=1}^{C} y_{ij} \log(p_{ij}) \quad \text{or} \quad \frac{1}{N} \sum_{i=1}^{N} \biggl(Y_{\text{actual}} - Y_{\text{predict}}\biggr)^2
\end{equation}

The first part corresponds to the loss function for a classification task, while the second part represents to the loss function for a regression task. $y_{ij}$ represents the true label for class $j$ of sample $i$. $p_{ij}$ is  the model's predicted probability for class $j$ of sample $i$.
\begin{equation}
\mathcal{L}_{\text{total}} = \mathcal{L}_{\text{modal x}} + \mathcal{L}_{\text{modal y}} + \mathcal{L}_{\text{task}}
\end{equation}
The overall loss function of the model combines these three losses, aiding in simultaneously optimizing the accuracy of completed modalities and the performance in the specific downstream task.

\subsection{Model Training and Evaluation}
We trained an LSTM-based network on datasets completed using different methods, including traditional interpolation, mean imputation, and our proposed MC-DBN. The performance was evaluated using RMSE, F1, and MAPE scores to assess the accuracy and predictive capability of each method.

\section{Experiment}
In this section, we elaborate on our experimental setup, comparing our MC-DBN methodology with other established approaches and conducting ablation experiments.

\subsection{Data Preparation}

\begin{figure}[ht]
    \centering
    \includegraphics[width=\linewidth]{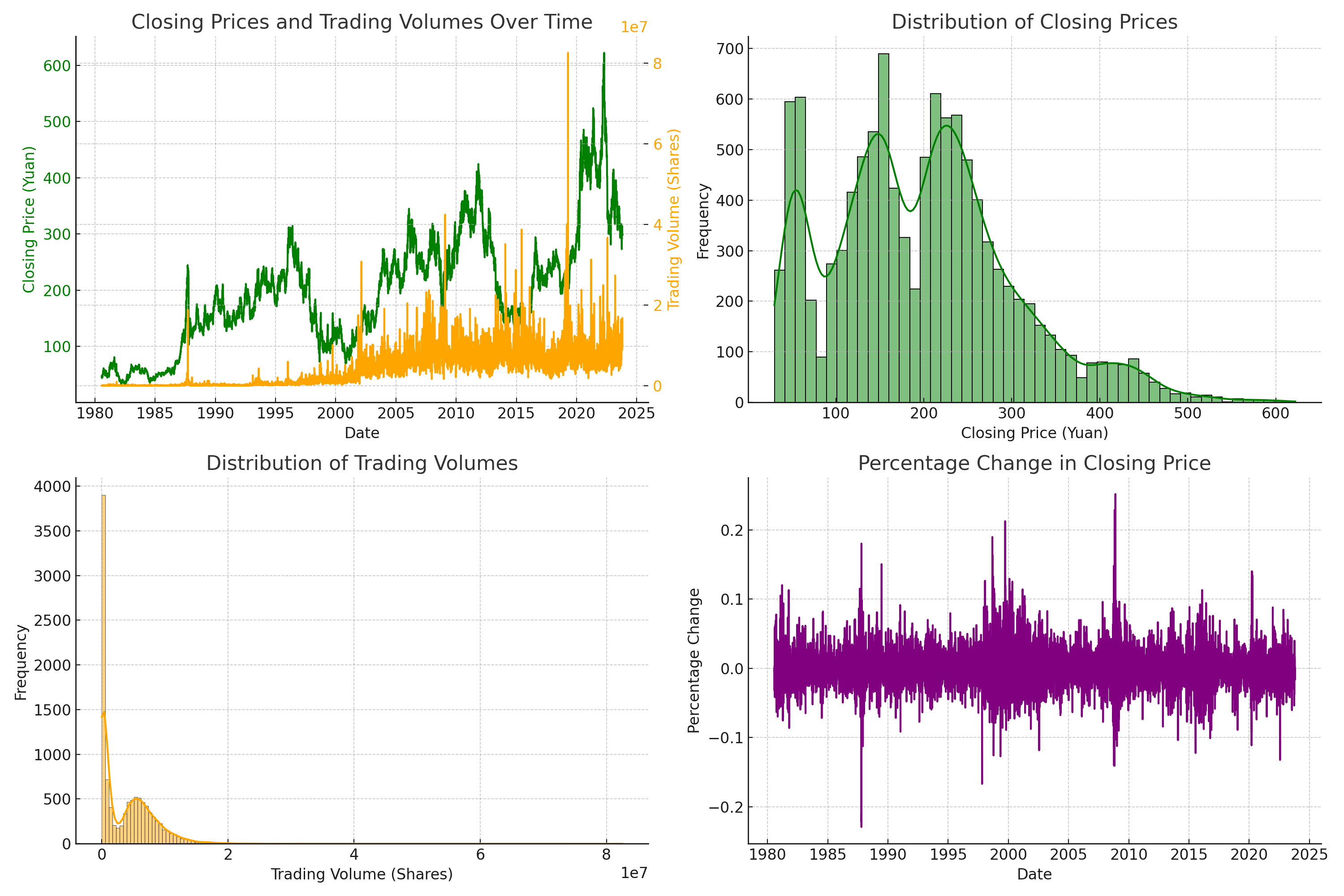}
    \caption{Combined Visualizations of NEWMONT (NEM.N) Stock Data}
      \label{stock}
\end{figure}

Our study encompasses two primary datasets: a financial market dataset and the esteemed MIT-BIH Arrhythmia Database. The financial market dataset comprises a collection of stock opening prices and associated discrete news events, covering the period from January 1, 2020, to January 1, 2023. This dataset has been meticulously preprocessed to meet the input criteria of our analytical models. A visualization of this stock data is provided in Figure \ref{stock}.

\begin{figure}[ht]
    \centering
    \includegraphics[width=\linewidth]{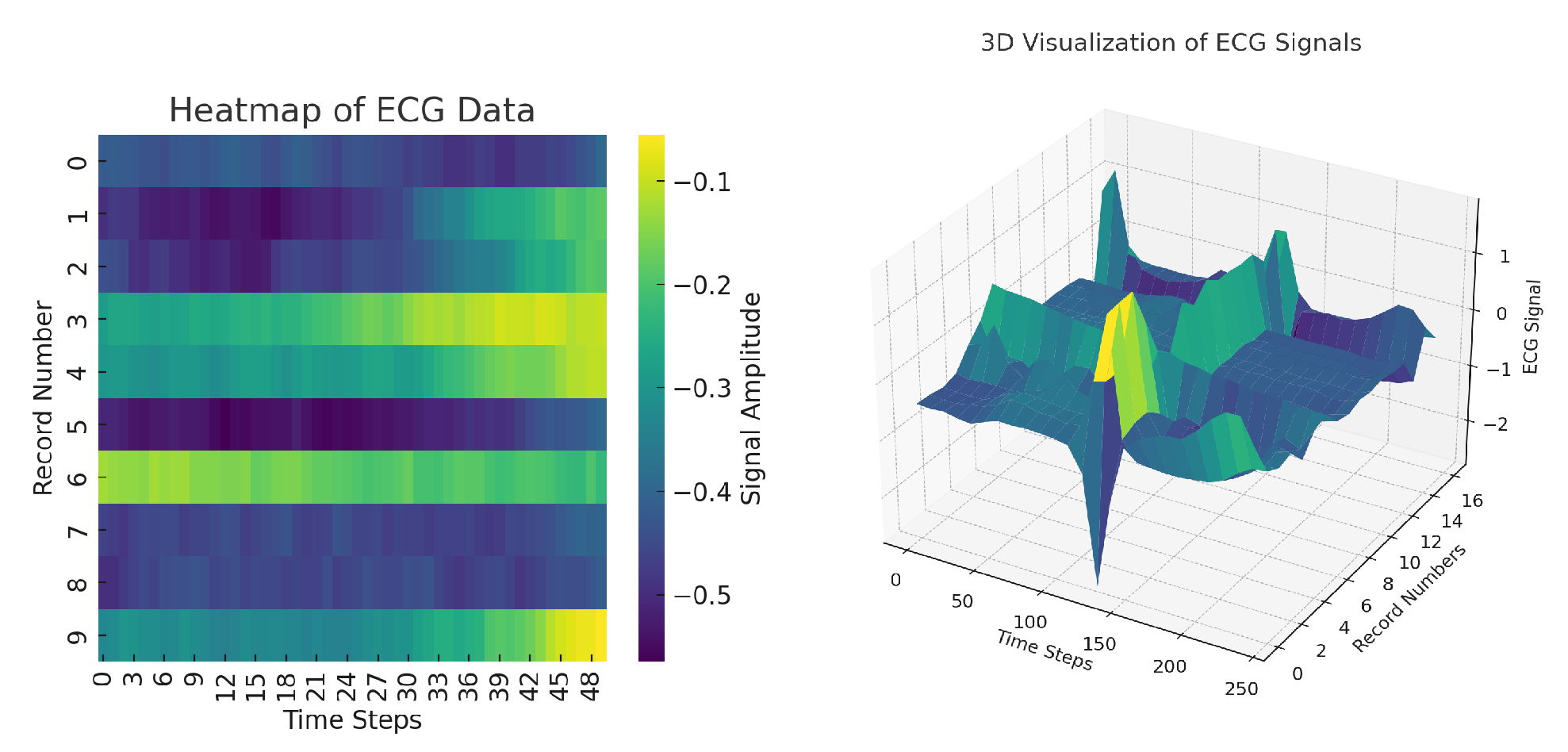}
    \caption{Combined Visualizations of MIT-BIH ECG Data}
    \vspace{-15pt}
     \label{mit}
\end{figure}

In the realm of cardiac research, the MIT-BIH Arrhythmia Database holds a distinguished position as a comprehensive repository for electrocardiogram (ECG) signals, instrumental in the development and validation of ECG heartbeat classification algorithms \cite{932724}. Curated jointly by the Massachusetts Institute of Technology and Beth Israel Hospital (now known as Beth Israel Deaconess Medical Center) in Boston, this dataset has been employed extensively for investigating cardiac arrhythmia detection and diagnostic methodologies. A visualization of the arrhythmia dataset is presented in Figure \ref{mit}.

\subsection{Comparative Experiment}
In our research, we integrated two diverse datasets: our unique stock data and the publicly available MIT-BIH arrhythmia dataset. This integration was essential to test the effectiveness and general applicability of our Deep Belief Network-based multimodal data completion method (MC-DBN).

This phase of our experiment focused on using various methods to complete a multimodal dataset. The dataset comprised two types of data: the opening prices of stocks in the medical sector and discontinuous news data, both covering the period from January 1, 2020, to January 1, 2023. The discontinuous news data were the primary target for completion using different methods. Subsequently, these datasets were utilized to train an LSTM-based network for predicting stock opening prices. We evaluated the network's performance using RMSE, MAPE, F1, and Accuracy Scores. RMSE and MAPE measured the model's error magnitude, whereas the F1 score assessed the model's accuracy in predicting price movements. The network's performance provided insights into the effectiveness of the multimodal data completion methods. To minimize error, we averaged the results from tests on ten different stocks.

As Table \ref{Compare1} illustrates, the single-modal data, being dependent on one data type for prediction, showed the highest RMSE and MAPE values, indicating larger prediction errors. In contrast, the use of multimodal data reduced these errors and improved price movement prediction accuracy. Notably, the MC-DBN processed multimodal data exhibited superior performance across all metrics, attributed to its ability to learn nonlinear features and ensure that the supplemented news data reflected actual news trends and variability.

The second Table \ref{Compare2} presents a similar comparison but focuses on the MIT-BIH dataset. As well, the MC-DBN method outperforms other approaches, reinforcing its efficacy in dealing with multimodal data for accurate predictions.

\begin{table}[ht]
\centering
\vspace{-5pt}
\caption{Comparison of other data completion methods on Stock dataset}
\vspace{-8pt}
\resizebox{0.5\textwidth}{!}{%
\begin{tabular}{lcccc}
\toprule
Method & \textbf{RMSE $\downarrow$}  & \textbf{MAPE $\downarrow$} & \textbf{F1 $\uparrow$} & \textbf{Accuracy $\uparrow$} \\
\midrule
Single modal data & 0.341 & 0.391 & 0.824 & 0.845\\
Multimodal data w/ Zero Filling & 0.295  & 0.387 & 0.792 & 0.798 \\
Multimodal data w/ Forward Fill  & 0.286  & 0.365 & 0.832 & 0.852 \\
Multimodal data w/ Mean Imputation & 0.279  & 0.361 & 0.856 & 0.870 \\
\midrule
\textbf{MC-DBN (ours)} & \textbf{0.268} & \textbf{0.346}   & \textbf{0.874} &  \textbf{0.913}\\
\bottomrule
\end{tabular}}
\vspace{-5pt}
\label{Compare1}
\end{table}

\begin{table}[ht]
\centering
\vspace{-5pt}
\caption{Comparison of other data completion methods on MIT-BIH dataset}
\vspace{-8pt}
\resizebox{0.5\textwidth}{!}{%
\begin{tabular}{lcc}
\toprule
Method  & \textbf{F1 $\uparrow$} & \textbf{Accuracy $\uparrow$} \\
\midrule
Single modal data &  0.852 & 0.874\\
Multimodal data w/ Zero Filling  & 0.867 & 0.886 \\
Multimodal data w/ Forward Fill  & 0.876 & 0.892 \\
 Multimodal data w/ Mean Imputation & 0.893 & 0.914 \\
 \midrule
\textbf{MC-DBN(ours)}   & \textbf{0.964} &  \textbf{0.982}\\
\bottomrule
\end{tabular}}
\vspace{-10pt}
\label{Compare2}
\end{table}


\subsection{Ablation Experiment}
In our ablation experiments, focusing on stock opening price prediction, we analyzed the roles of the decoder and two loss functions within the MC-DBN. Table \ref{table:loss_ablation} demonstrates that $\mathcal{L}_{\text{modal x}}$ and $\mathcal{L}_{\text{modal y}}$  complement each other, thereby enhancing the completion capability of the MC-DBN. Furthermore, $\mathcal{L}_{\text{modal x}}$ is crucial in guiding new data completion during model training, significantly improving model accuracy and consistency. Table \ref{table:ablation} underscores the importance of employing LSTM for decoding news data and Transformers for sequential data decoding, highlighting their superior performance over traditional linear decoding methods.

\begin{table}[ht]
\centering
\vspace{-10pt}
\caption{Ablation Experiment about Loss}
\vspace{-8pt}
\begin{tabular}{lcccc}
\toprule
 & \textbf{RMSE $\downarrow$}  & \textbf{MAPE $\downarrow$}  & \textbf{F1 $\uparrow$} & \textbf{Accuracy $\uparrow$}\\
    \midrule
    $\mathcal{L}_{\text{modal y}} $  & 0.282  &0.358  & 0.853 & 0.869\\
    $\mathcal{L}_{\text{modal x}} $  & 0.271 &0.349   & 0.872 & 0.873\\
    $\mathcal{L}_{\text{modal x}} + \mathcal{L}_{\text{modal y}} $  & \textbf{0.268}&\textbf{0.346} & \textbf{0.874} & \textbf{0.913} \\
    \bottomrule
\end{tabular}
\vspace{-5pt}
\label{table:loss_ablation}
\end{table}

\begin{table}[ht]
\centering
\vspace{-10pt}
\caption{Ablation Experiment about core components of decoder}
\vspace{-5pt}
\begin{tabular}{lcccc}
\toprule
 & \textbf{RMSE $\downarrow$}  & \textbf{MAPE $\downarrow$} & \textbf{F1 $\uparrow$} & \textbf{Accuracy $\uparrow$}  \\
    \midrule
    Both Linear& 0.276 & 0.358  & 0.854 & 0.862  \\
    Only Transformer & 0.274   &0.352  & 0.859 & 0.875 \\
    Only LSTM & 0.272  &0.349   & 0.867 & 0.874 \\
    LSTM + Transformer & \textbf{0.268}  &\textbf{0.346} & \textbf{0.874} & \textbf{0.913} \\
    \bottomrule
\end{tabular}
\vspace{-5pt}
\label{table:ablation}
\end{table}

\section{Conclusion}

This research has made significant advancements in the field of multimodal data analysis, particularly in stock market forecasting and heart rate monitoring. We utilized Deep Belief Networks (DBN) and attention mechanisms to propose an innovative modal completion mechanism, effectively addressing issues of data missingness and long-term dependencies in stock market prediction and heart rate monitoring. This mechanism includes an advanced encoder-decoder framework that integrates multimodal information efficiently through multi-head cross-attention mechanisms and selective adoption of target feature extraction architectures. Additionally, we designed two types of loss functions to optimize data completion quality and enhance network performance. Empirical tests demonstrate that our model exhibits high accuracy and predictive performance in complex data environments, offering new tools for both theoretical and practical applications in the field of multimodal data analysis. In the future, this approach is expected to be applicable to a wider range of fields where multimodal data is crucial.

\newpage

\bibliographystyle{elsarticle-num} 
\bibliography{ref}

\end{document}